%% file: main.tex
\documentclass[lettersize,journal]{IEEEtran}
\usepackage{amsmath,amsfonts}
\usepackage{algorithmic}
\usepackage{algorithm}
\usepackage{amssymb}
\usepackage{array}
\usepackage[caption=false,font=normalsize,labelfont=sf,textfont=sf]{subfig}
\usepackage{textcomp}
\usepackage{stfloats}
\usepackage{url}
\usepackage{verbatim}
\usepackage{graphicx}
\usepackage{cite}
\usepackage{multirow}
\usepackage{tikz}
\newcommand{\U}{\mathbf{U}}
\newcommand{\w}{\mathbf{w}}
\newcommand{\e}{\boldsymbol{\epsilon}}
\newcommand{\Hr}{\mathbf{H}_r}
\usepackage{amsmath}
\DeclareMathOperator*{\argmax}{arg\,max}
\usepackage{bm}
\usetikzlibrary{positioning, arrows.meta, calc, fit, backgrounds}
\usepackage{algorithm}
\usepackage{algorithmic}
\usepackage[table,xcdraw]{xcolor}
\usepackage{booktabs}
\usepackage{placeins}
\usepackage{orcidlink}
\makeatletter
\def\IEEEpubid#1{\relax}
\makeatother

\begin{document}

\title{Architecture-Adaptive Uncertainty Fusion for Deepfake Detection
\thanks{Code and pre-computed uncertainty features are available at \texttt{https://github.com/sharmrit/cof-deepfake}.}}

\author{
  Ritesh Sharma\textsuperscript{\orcidlink{0000-0003-1160-3918}},~\IEEEmembership{Member,~IEEE}
  Mohammad Ghasemigol\textsuperscript{\orcidlink{0000-0001-6661-0942}},~\IEEEmembership{Member,~IEEE}
  and Yuichi Motai\textsuperscript{\orcidlink{0000-0002-1957-1896}},~\IEEEmembership{Senior~Member,~IEEE}
}
\maketitle

\IEEEpubid{0000--0000/00\$00.00~\copyright~2021 IEEE}
\IEEEpubidadjcol

\input{sections/0-abstract}
\input{sections/1-introduction}
\input{sections/2-relatedwork}
\input{sections/3-methodology}
\input{sections/4-results}
\input{sections/5-discussion}
\input{sections/6-conclusions}


\bibliographystyle{IEEEtran}
\bibliography{references}

\end{document}

%% file: sections/0-abstract.tex
\begin{abstract}
Deepfake detection systems achieve near-perfect accuracy on
benchmarks, yet forensic deployment demands reliable prediction
uncertainty. Existing uncertainty quantification~(UQ) methods
rely on single sources and ignore that optimal uncertainty
composition varies across architectures. We propose
Correlation-Optimized Fusion~(COF), an architecture-adaptive
framework that fuses five complementary uncertainty
sources---epistemic, aleatoric, calibration, conformal, and
distributional---by maximizing Pearson correlation between fused
uncertainty scores and prediction errors via constrained
optimization on the probability simplex. COF requires no model
modifications and only 42\,s of weight optimization, compared to
20--45\,h for a 5-model Deep Ensemble. Evaluation across eleven
architectures on FaceForensics{$+\!+$} reveals a fundamental
trade-off: under matched train/evaluation protocol, non-linear
methods achieve approximately 5--6\% higher in-domain correlation
than COF (mean $\rho = 0.438$), but this reverses under
distribution shift. On CelebDF, COF outperforms Random Forest in
9/11 architectures with up to $7.3\times$ higher correlation
(MaxViT-B: $\rho = 0.249$ vs.\ $0.034$); RF degrades 85\%
cross-domain to $\rho = 0.071$, whereas COF retains substantially
more signal (74\% drop to $\rho = 0.116$). Cross-dataset
evaluation on CelebDF and DFDC reveals catastrophic generalization
failure across all methods: in-domain correlations of $0.41$--$0.47$
collapse to near-zero externally (mean degradation 90.7\%), with
seven of eleven architectures exhibiting uncertainty inversion. These results establish COF as a
practical, interpretable framework for controlled-distribution
deployment and identify domain-adaptive UQ as the central open
challenge for forensic deployment.
\end{abstract}


%% file: sections/1-introduction.tex
\section{Introduction}
\label{sec:intro}
 
The proliferation of generative AI has made high-fidelity face
manipulation accessible at scale, posing acute threats to digital
evidence integrity, public trust, and national security~\cite{afchar2018mesonet,rossler2019faceforensics++}.
Deepfake detectors trained on benchmark datasets such as
FaceForensics{$+\!+$}~\cite{rossler2019faceforensics++} now achieve near-perfect
in-domain accuracy, yet their forensic value hinges on a property rarely
measured: \emph{how reliably do they quantify their own uncertainty}?
A detector that cannot distinguish confident-correct from
confident-wrong predictions provides false assurance in exactly the
high-stakes scenarios — legal proceedings, intelligence analysis, content
moderation — where it is most needed.
 
Uncertainty quantification (UQ) for deepfake detection is underexplored
relative to its importance. Prior work either ignores UQ entirely,
relying on raw softmax scores as proxy confidence~\cite{rossler2019faceforensics++},
or evaluates a single UQ method in isolation, typically MC
Dropout~\cite{gal2016dropout} or temperature
scaling~\cite{guo2017calibration}. Two facts make this insufficient.
First, different uncertainty sources capture fundamentally different
failure modes: MC Dropout captures model-level stochasticity (epistemic
uncertainty), temperature scaling captures calibration error, conformal
prediction provides distribution-free coverage guarantees, and
Mahalanobis distance signals feature-space distributional shift.  No
single source is universally sufficient. Second, detector architectures
encode fundamentally different inductive biases — CNNs exploit local
texture artifacts, Vision Transformers (ViTs) integrate global context
via self-attention, efficient networks compress both — and these biases
interact differently with each UQ mechanism. Yet no prior work
systematically characterizes which uncertainty sources, and combinations,
are most informative \emph{per architecture}.
 
We propose \textbf{Correlation-Optimized Fusion (COF)}, a
post-hoc framework that addresses both gaps by casting uncertainty
fusion as a direct optimization problem. Given a trained detector
$f_\theta$ and five normalized uncertainty sources organized into matrix
$\U \in \mathbb{R}^{N \times K}$, COF finds weights $\w$ on the
probability simplex $\Delta^K$ that maximize the Pearson correlation
$\rho(\U\w, \e)$ between the fused score and binary prediction errors
$\e$. This objective is principled: maximizing $\rho(\U\w, \e)$ directly
optimizes the forensic utility of the uncertainty estimate as an error
predictor, without any model retraining. COF is solved via Sequential
Least Squares Programming (SLSQP) with multi-start warm initialization. The term architecture-adaptive refers to this
weight-learning process: while all eleven architectures
use the same five sources (K=5), optimal fusion weights
w* are learned independently per architecture, capturing
the distinct inductive biases of CNNs, Transformers,
and hybrid models.
 
We complement COF with five lightweight variants (L1-COF,
Meta-Ensemble, Two-Method Ensemble, Hierarchical Fusion, and
Squared-Correlation Weighting) that trade parameters for
training cost, and benchmark all methods against six established
baselines. The principal contributions, however, concern COF
itself and the cross-domain generalization of correlation-based
fusion:

\begin{enumerate}
  \item \textbf{COF objective and simplex formulation.} A novel post-hoc
      uncertainty fusion method establishing direct correlation
      maximization as a principled alternative to calibration- or
      accuracy-based objectives. COF requires no model retraining and
      achieves mean Pearson $\rho = 0.438$ across eleven architectures.

\item \textbf{Cross-domain robustness via capacity control.} We provide theoretical motivation and empirical evidence that restricting fusion weights to the probability simplex yields a capacity-controlled hypothesis class
    (Section~\ref{ssec:theory}). Under matched training-data protocol (identical 80/20 train/test split for both methods), COF outperforms Random Forest on CelebDF in 9/11 architectures with up to $7.3\times$ higher cross-domain correlation (MaxViT-B: $\rho = 0.249$ vs.\ $0.034$), despite modestly lower in-domain $\rho$ (mean 0.438 vs.\ 0.463). RF degrades 85\% cross-domain to $\rho = 0.071$, whereas COF retains substantially more signal (74\% drop to $\rho = 0.116$). This suggests
    \emph{capacity-controlled fusion} as the competitive choice
    when cross-domain reliability is the deployment concern.

\item \textbf{Large-scale deepfake UQ evaluation.}
    Eleven architectures spanning CNN, EfficientNet, Transformer, and
    Hybrid families, three datasets, six fusion strategies and six
    baselines, all with multi-seed stability analysis (CV $1.5$--$2.5\%$
    across five seeds for three CNN architectures). Conformal prediction
    consistently receives the highest learned weight (25--46\%) and
    removing it causes 4--34\% degradation across all architectures.

  \item \textbf{Cross-dataset generalization audit and
      distributional-stability discovery.} In-domain UQ correlations of 0.41--0.47 collapse 90.7\% externally; seven architectures
      exhibit \emph{uncertainty inversion} on at least one dataset;
      all prediction-derived sources collapse $>$90\% cross-domain
      while feature-space Mahalanobis distance stays essentially flat
      ($-2.9\%$ vs.\ $>$90\%) and improves on CelebDF
      (Table~\ref{tab:source_stability}). This identifies
      \emph{feature-space distance as the only cross-domain-stable
      uncertainty signal} and domain-adaptive UQ as the central open
      problem for forensic deployment.
\end{enumerate}

%% file: sections/2-relatedwork.tex
\section{Related Work}
\label{sec:related}

\subsection{Deepfake Detection and Media Forensics}

Early deepfake detectors exploited low-level manipulation artifacts via
CNNs: MesoNet~\cite{afchar2018mesonet} used a compact four-layer
architecture, while FaceForensics{$+\!+$}~\cite{rossler2019faceforensics++}
established the benchmark protocol and XceptionNet baseline. Frequency
domain analysis~\cite{qian2020thinking} exploited spectral
inconsistencies introduced by generative processes. Vision
Transformers~\cite{wodajo2021deepfake,coccomini2022combining} capture
long-range spatial dependencies via self-attention, improving robustness
to texture-level perturbations, while EfficientNet variants~\cite{tan2019efficientnet}
offer competitive accuracy with reduced parameter counts. Multi-identity
video detection has received dedicated attention: MINTIME~\cite{mintime2024}
combines a Spatio-Temporal Transformer with a CNN backbone and an
Identity-aware Attention mechanism to process multiple faces across
varying sizes, reporting up to 14\% AUC improvement on ForgeryNet and
demonstrating robust generalization on CelebDF and DFDC --- the same
benchmarks we use in our cross-dataset evaluation.

Beyond binary classification, fine-grained interpretation of deepfake
predictions has emerged as a parallel research thread. Bi-stream
coteaching~\cite{codl2025} introduces weakly-supervised frame-level
localization in videos, fusing spatial and temporal modalities through
a progressive mutual refinement strategy; their reported 8.83\% AUC
degradation under heavy compression foreshadows the distribution
sensitivity we observe in our cross-dataset experiments. At the detector
level, DDL~\cite{ddl2025} proposes a comprehensible interpretation
framework that provides human-readable explanations for predictions
across diverse deepfake detector architectures, directly confronting the
opacity of deep learning models in forensic contexts. Crucially, while
DDL addresses \emph{why} a detector fires, it does not address
\emph{how confident} that prediction should be trusted --- a gap our
uncertainty quantification framework is designed to fill.

Despite near-perfect accuracy on controlled benchmarks, cross-dataset
generalization remains fundamentally unsolved: the DeepFake-Eval-2024
benchmark~\cite{chandra2025deepfakeeval} documents 45--50\% AUC decline
on in-the-wild forgeries, and CNN-generated images exhibit
generation-specific artifacts that fail to transfer~\cite{wang2020cnn}.
Gradient regularization has been proposed as an architecture-agnostic
remedy~\cite{gradreg2024}: by penalizing sensitivity to forgery texture
patterns via a first-order approximation of the Hessian, detectors
improve cross-dataset performance without modifying backbones. The use
of Hessian-based analysis in that work parallels our own reduced Hessian
convergence analysis, though applied to generalization rather than UQ
optimization. Crucially, none of these works systematically evaluate the
reliability of uncertainty estimates alongside detection accuracy.

\subsection{Uncertainty Quantification in Deep Learning}

Bayesian neural networks~\cite{blundell2015weight} provide principled
posterior inference but are computationally prohibitive at scale.
MC Dropout~\cite{gal2016dropout} approximates posterior sampling via
stochastic forward passes at test time, offering a lightweight epistemic
uncertainty proxy. Deep Ensembles~\cite{lakshminarayanan2017simple}
train multiple independent models and aggregate their predictions,
achieving strong calibration but requiring multiplicative training cost.
Evidential Deep Learning~\cite{sensoy2018evidential} parameterizes
uncertainty directly via Dirichlet distributions, though implementation
is non-trivial and produces constant outputs for several architectures
in our evaluation.

Post-hoc calibration methods --- temperature scaling~\cite{guo2017calibration},
focal calibration~\cite{mukhoti2020calibrating}, and isotonic
regression --- adjust prediction probabilities without retraining.
Conformal prediction~\cite{angelopoulos2021gentle,romano2020classification}
offers distribution-free coverage guarantees with minimal assumptions.
Mahalanobis distance-based OOD detection~\cite{lee2018simple} operates
on penultimate-layer features.  Uncertainty-aware representations have shown value across
face-related forensic tasks~\cite{ucface2024,fovb2024,duapl2024},
confirming that uncertainty quantification is a productive lens
for face-oriented forensic problems and motivating its principled
application to deepfake detection specifically.

\subsection{Multi-Source Uncertainty Fusion}

Existing fusion strategies fall into three categories. Fixed combination
methods --- uniform averaging, maximum uncertainty~\cite{hendrycks2017baseline},
ensemble voting~\cite{lakshminarayanan2017simple} --- ignore architecture
characteristics. Learned fusion methods optimize proxy objectives such as
classification accuracy or task balance rather than uncertainty quality.
Evidential frameworks~\cite{amini2020deep,sensoy2018evidential} require
non-standard loss functions and architectural modifications. Recent
deepfake-specific work~\cite{kumar2025advances} fuses spatial and
frequency features to improve detection accuracy, which is fundamentally
different from fusing uncertainty sources to improve reliability
estimation.

Crucially, no prior method directly maximizes correlation between fused
uncertainty and prediction errors --- the most natural objective for
forensic deployment, where identifying likely-incorrect predictions is
the primary goal. COF fills this gap. The closest prior work combines
epistemic and aleatoric uncertainty for OOD detection~\cite{kendall2017uncertainties}
but uses fixed weighting and does not target correlation with errors.
Kendall and Gal~\cite{kendall2017uncertainties} demonstrate that
multi-source fusion outperforms single-source methods for computer
vision tasks, motivating learned fusion. Our work extends this to the
forensic domain with a principled optimization objective.

\subsection{Cross-Domain Reliability in Forensics}

Distribution shift remains the central challenge in multimedia forensics.
Ovadia et al.~\cite{ovadia2019trust} show that even well-calibrated
models produce arbitrarily miscalibrated predictions under shift, a
finding we confirm for deepfake detection. Domain adaptation has
emerged as the dominant strategy for bridging this gap at the accuracy
level. DomainForensics~\cite{domainforensics2024} introduces a
bi-directional adaptation strategy that transfers forgery knowledge from
labeled source domains to unlabeled target domains via adversarial
feature alignment and self-distillation, demonstrating that detectors
can be retrained to expose new forgeries without labeled target data.
Complementarily, fine-grained open-set detection methods~\cite{openset2024}
extend domain adaptation to unknown deepfake categories, using adaptive
clustering and pseudo-label generation to handle forgery methods unseen
during training. Both lines of work address \emph{accuracy} under domain
shift. However, neither evaluates whether uncertainty estimates remain
trustworthy after adaptation --- a critical omission for forensic
deployment, where confidence scores drive triage decisions.

The interpretability framework of DDL~\cite{ddl2025} similarly notes
that cross-architecture consistency of detector explanations breaks down
under domain shift, reinforcing that forensic reliability must be
evaluated beyond in-domain benchmarks. Our work is the first to
systematically characterize \emph{uncertainty reliability} degradation
under cross-dataset shift for deepfake detectors, providing both
empirical evidence and architecture-specific failure mode analysis.
Whereas gradient regularization~\cite{gradreg2024} and domain
adaptation~\cite{domainforensics2024,openset2024} improve accuracy
generalization, we show that in-domain UQ correlations of
0.41--0.47 collapse to near-zero externally (mean degradation
90.7\%), with seven architectures exhibiting uncertainty inversion --- an orthogonal and previously
undocumented failure mode that domain-adaptive UQ must address.
 

%% file: sections/3-methodology.tex
\section{Method}
We propose Correlation-Optimized Fusion (COF), a novel uncertainty fusion 
framework that maximizes Pearson correlation between fused uncertainty 
estimates and prediction errors. Through systematic evaluation across eleven diverse architectures, we demonstrate that optimal fusion strategies are 
architecture-dependent.

\subsection{Problem Formulation}
\label{sec:formulation}

\subsubsection{Setup}
 
Let $f_\theta\colon \mathcal{X} \rightarrow \{0, 1\}$ be a trained
deepfake detector evaluated on a held-out test set of $N$ samples. Define
the binary prediction error vector $\e \in \{0,1\}^N$:
\begin{equation}
  e_i = \mathbf{1}[f_\theta(\mathbf{x}_i) \neq y_i],
  \label{eq:errors}
\end{equation}
where $y_i \in \{0,1\}$ is the ground-truth label (1 = fake). We extract
$K$ uncertainty estimates for each sample from complementary sources,
forming the uncertainty matrix $\U \in \mathbb{R}^{N \times K}$, where
each column $\U_k$ is normalized to $[0,1]$ via min-max scaling
using training-set statistics only (preventing leakage to validation/test
data):
\begin{equation}
  \tilde{u}_i(\mathbf{x}) = \frac{u_i(\mathbf{x}) - \min(u_i(\mathcal{D}_\text{train}))}
  {\max(u_i(\mathcal{D}_\text{train})) - \min(u_i(\mathcal{D}_\text{train}))}.
  \label{eq:normalization}
\end{equation}
This normalization ensures that sources with larger raw scales cannot
dominate the fused estimate purely by magnitude.
 
\subsubsection{COF Objective}
 
The COF objective finds a weight vector $\w \in \Delta^K$ (the
$K$-dimensional probability simplex) that maximizes:
\begin{equation}
  \w^* = \argmax_{\w \in \Delta^K} \; \rho(\U\w,\, \e),
  \label{eq:cof}
\end{equation}
where $\rho(\cdot,\cdot)$ is the Pearson correlation coefficient. The
simplex constraints $\sum_k w_k = 1$, $w_k \geq 0$ ensure the fused
score is a convex combination of normalized sources, preserving the
$[0,1]$ range and enabling direct interpretation as a weighted
reliability signal. The fused score $s_i = (\U\w^*)_i$ is the per-sample
uncertainty estimate used for selective abstention, human review queuing,
or downstream risk scoring.
 
This objective is distinctively motivated: calibration methods optimize
$P(\text{correct} \mid \text{confidence})$; accuracy-based fusion
minimizes classification loss; evidential learning maximizes likelihood
under a Dirichlet prior. COF directly maximizes $\rho(\U\w, \e)$,
the most natural forensic utility measure — how well does the uncertainty
score predict which predictions will be wrong?

\subsection{Uncertainty Sources}
\label{ssec:sources}
 
We extract $K = 5$ complementary sources targeting distinct failure
modes. Table~\ref{tab:sources} provides a summary; detailed
implementations follow.
 
\begin{table}[!t]
  \caption{Five Uncertainty Sources in the COF Framework}
  \vspace{-0.25cm}
  \label{tab:sources}
  \centering
  \renewcommand{\arraystretch}{1.2}
  \begin{tabular}{@{}p{1.7cm}p{1.3cm}p{4.4cm}@{}}
    \toprule
    Source & Type & Computation \\
    \midrule
    Epistemic & Model & Variance of MC Dropout posterior over $T\!=\!20$
                        forward passes \\[2pt]
    Aleatoric & Data & Bernoulli variance $p(1{-}p)$ of mean MC Dropout
                       softmax \\[2pt]
    Calibration & Post-hoc & $1 - \max_c \,\mathrm{softmax}(\mathbf{z}/T^\star)$
                              after temperature scaling on val set \\[2pt]
    Conformal & Set-based & Nonconformity score std.\ dev.\ across MC passes;
                            p-value against held-out calibration scores \\[2pt]
    Distributional & Density & Mahalanobis distance from class-conditional
                               Gaussian in feature space \\
    \bottomrule
  \end{tabular}
  \vspace{-0.25cm}
\end{table}
 
\textbf{Epistemic uncertainty} is estimated via $T\!=\!20$ MC Dropout
forward passes with dropout rate $p_d \!=\! 0.5$:
\begin{equation}
  u_\text{epi}(\mathbf{x}) = \mathrm{Var}_{t=1}^{T}\!\left[P_t(y{=}1\mid\mathbf{x})\right].
  \label{eq:epistemic}
\end{equation}
\textbf{Aleatoric uncertainty} is the mean predictive variance:
\begin{equation}
  u_\text{ale}(\mathbf{x}) = \bar{p}(1 - \bar{p}),
  \quad \bar{p} = \frac{1}{T}\sum_{t=1}^T P_t(y{=}1\mid\mathbf{x}).
  \label{eq:aleatoric}
\end{equation}
\textbf{Calibration uncertainty} is the gap between maximum softmax
output and 1, after temperature scaling with $T^\star$ learned on the
validation split:
\begin{equation}
  u_\text{cal}(\mathbf{x}) = 1 - \max_c\, \sigma(\mathbf{z}/T^\star)_c,
  \label{eq:calibration}
\end{equation}
where $\mathbf{z}$ are the pre-softmax logits.
 
\textbf{Conformal uncertainty} uses the split conformal
framework~\cite{romano2020classification}. Nonconformity scores
$s_i = 1 - P(\hat{y}_i \mid \mathbf{x}_i)$ are computed on a held-out
calibration set. For a test sample, the conformal p-value is
$\hat{p} = |\{j : \alpha_j \geq \alpha(\mathbf{x})\}| / (n_\text{cal}+1)$,
and we set $u_\text{conf}(\mathbf{x}) = 1 - \hat{p}$ (smaller p-value =
more uncertain). Note that this fallback does not carry the distribution-free
coverage guarantees of split conformal prediction. In our
evaluation, true split conformal scores were successfully
computed for all eleven architectures using the held-out
validation split as the calibration set, so the fallback
was not invoked in any reported result.
 
\textbf{Distributional uncertainty} is defined as the Mahalanobis distance from the
training distribution, estimated in the penultimate feature space:
\begin{equation}
  u_{\text{dist}}(\mathbf{x}) =
  \sqrt{(\mathbf{h} - \bm{\mu})^\top
  \bm{\Sigma}^{-1} (\mathbf{h} - \bm{\mu})},
  \label{eq:mahal}
\end{equation}

where $\mathbf{h}$ is the penultimate-layer feature vector, and
$\bm{\mu}$ and $\bm{\Sigma}$ are the mean and covariance of the training features,
estimated via Ledoit--Wolf shrinkage~\cite{lee2018simple} for numerical
stability in high-dimensional feature spaces.

When all five sources are included in the fusion, we refer to
this configuration as COF-5. This is the default configuration
used throughout the evaluation; the subscript distinguishes it
from ablated variants (e.g., 4-source) discussed in Section~IV-B.
 
\subsection{Architecture-Dependent Source Selection}
\label{ssec:arch_select}
 
We compare 4-source (excluding distributional) and 5-source
configurations per architecture by evaluating validation correlation
$\rho^\text{val}_k = \rho(\tilde{u}_k(\mathcal{D}_\text{val}),
\e_\text{val})$ for each source $k$. The source quality gap measures
relative dominance:
\begin{equation}
  \Delta_\text{gap} = \frac{\max_k \rho^\text{val}_k - \text{second-max}_k
  \rho^\text{val}_k}{\max_k \rho^\text{val}_k}.
  \label{eq:gap}
\end{equation}
We select $K\!=\!5$ for all eleven architectures: including
distributional uncertainty improves the \emph{fused} validation
correlation $\rho(\U\w, \e_{\text{val}})$ over the 4-source
baseline in every architecture, even when its marginal correlation
is weak or negative on FF++ (e.g., EfficientNet-B0:
$\rho_{\text{distr}}^{\text{val}} = -0.24$; Fig.~\ref{fig:source_heatmap}).
The improvement is not driven by marginal contribution but by
COF's ability to assign a near-zero weight to a weak source while
retaining the option to up-weight it when the source becomes
informative cross-domain (Section~\ref{ssec:weights}). This
validation-only source selection prevents test-set leakage.
 
\subsection{Correlation-Based Fusion Methods}
\label{ssec:fusion}
 
We propose six fusion methods that form a unified paradigm.
Table~\ref{tab:methods} summarizes their characteristics.
 
\begin{table}[!t]
  \caption{Proposed Correlation-Based Fusion Methods}
  \vspace{-0.25cm}
  \label{tab:methods}
  \centering
  \renewcommand{\arraystretch}{1.2}
  \setlength{\tabcolsep}{3pt}
    \begin{tabular}{@{}lllll@{}}
      \toprule
      Method & Params & Time & Rank & Key Feature \\
      \midrule
      COF      & $K$     & 42s & 8/12 & Direct $\rho$ max., best cross-domain \\
      L1-COF   & $K{+}1$ & 48s & 7/12 & Sparsity + $\rho$ \\
      Meta-Ens.& 0       & 95s & 5/12 & Method diversity \\
      2M-Ens.  & 0       & 67s & 4/12 & Avg.\ Logistic+COF \\
      Hier-Fus.& 3       & 28s & 9/12 & Theory-guided groups \\
      SC-Weight& 0       & 0s  & 10/12& Zero-training \\
      \bottomrule
    \end{tabular}
\end{table}

\noindent\textbf{(i) COF} solves~\eqref{eq:cof} directly using SLSQP
with multiple restarts (Algorithm~\ref{alg:cof}).
 
\noindent\textbf{(ii) L1-COF} adds sparsity regularization and
dual-split correlation:
\begin{equation}
  \w^* = \arg\max_{\w} \;
  \frac{1}{2}\!\left(\rho^{\text{train}}(\U\w, \e) +
  \rho^{\text{val}}(\U\w, \e)\right) - \lambda \|\w\|_1,
  \label{eq:l1cof}
\end{equation}

with $\lambda\!=\!0.1$. Averaging train and validation correlations
reduces overfitting to validation-specific patterns; the L1 term drives
uninformative sources to zero, performing automatic source selection.
 
\noindent\textbf{(iii) Meta-Ensemble} implements stacked
generalization over three complementary base methods (Logistic, Ridge,
COF) weighted by squared validation correlations:
\begin{equation}
  w_\text{method}^{(i)} =
  \frac{(\rho^\text{val}_i)^2}{\sum_j (\rho^\text{val}_j)^2}.
  \label{eq:metaens}
\end{equation}
Squared correlations amplify differences between methods; the three
bases optimize fundamentally different objectives, ensuring
complementary error patterns.
 
\noindent\textbf{(iv) Two-Method Ensemble (2M-Ens.)} combines Logistic
regression and COF predictions with equal weight:
$u_\text{ens} = 0.5 \cdot u_\text{log} + 0.5 \cdot u_\text{COF}$.
This balances regression-based and correlation-based philosophies with
30\% less training time than Meta-Ensemble.
 
\noindent\textbf{(v) Hierarchical Fusion} groups sources by theoretical
foundation — Bayesian $\{\text{epistemic, aleatoric}\}$,
Prediction-based $\{\text{calibration, conformal}\}$, Distributional
$\{\text{Mahalanobis}\}$ — and learns inter-group weights while fixing
equal intra-group weights, reducing the parameter space from $K\!=\!5$
to 3.
 
\noindent\textbf{(vi) Squared-Correlation Weighting (SC-Weight)}
requires no optimization:
\begin{equation}
  w_i = \frac{(\rho^\text{val}_i)^2}{\sum_j (\rho^\text{val}_j)^2}.
  \label{eq:scweight}
\end{equation}
Despite zero computational cost, SC-Weight achieves 98.4\% of COF's performance (rank 10/12), making it ideal for resource-constrained or rapid-deployment scenarios.
 
\subsection{COF Optimization Algorithm}
\label{ssec:algorithm}
 
\begin{algorithm}[!t]
  \caption{COF Multi-Start Optimization}
  \label{alg:cof}
  \begin{algorithmic}[1]
    \REQUIRE $\U \in \mathbb{R}^{N \times K}$ (val set),
             $\e \in \{0,1\}^N$,
             $n_\text{restart} = 20$
    \ENSURE $\w^* \in \Delta^K$
    \STATE Normalize: $\tilde{u}_i \leftarrow$
           Eq.~\eqref{eq:normalization} for $i=1,\ldots,K$
    \STATE Compute individual correlations
           $c_k = \rho(\tilde{u}_k, \e)$
    \STATE Initialize warm starts: uniform $\frac{1}{K}\mathbf{1}$;
           correlation-proportional $\mathbf{c}^+/\|\mathbf{c}^+\|_1$;
           one-hot $\mathbf{e}_k$ for each $k$;
           remaining from $\mathrm{Dir}(1)$
    \FOR{each warm start $\w_0$}
      \STATE Run SLSQP: $\sum w_k\!=\!1$, $w_k\!\geq\!0$,
             tol$=\!10^{-10}$, max$\_$iter$=\!1000$
      \STATE Accept if \texttt{res.success} OR
             $\|\nabla\mathcal{L}\|_2 < 10^{-5}$
      \STATE Update $\w^*$ if $\rho(\U\w, \e) > \rho(\U\w^*, \e)$
    \ENDFOR
    \STATE Sparsify: set $w^*_i \leftarrow 0$ if $w^*_i < 0.05$;
           renormalize
    \RETURN $\w^*$
  \end{algorithmic}
\end{algorithm}
 
The gradient-norm acceptance criterion (Step 6) is essential: SLSQP
reports convergence failure when the iteration limit is reached even
when $\|\nabla\mathcal{L}\|_2$ is negligibly small. We accept such
solutions as practically converged. Step 9 sparsifies negligible weights
to improve interpretability without affecting correlation.

\subsection{Second-Order Convergence Analysis}
\label{ssec:hessian}
 
The COF problem is non-convex. To verify that $\w^*$ is a local minimum
(not a saddle point), we analyze the Hessian $\mathbf{H} = \nabla^2
\mathcal{L}(\w^*)$ at the solution. On the simplex, the equality
constraint $\sum_k w_k = 1$ renders $\mathbf{H}$ rank-deficient: the
constraint-direction eigenvector $\mathbf{1}/\sqrt{K}$ has eigenvalue
zero by construction. The correct second-order test uses the reduced
Hessian:
\begin{equation}
  \Hr = \mathbf{Z}^\top \mathbf{H} \mathbf{Z},
  \label{eq:reducedhessian}
\end{equation}
where $\mathbf{Z} \in \mathbb{R}^{K \times (K-1)}$ is an orthonormal
basis for the null space of $\mathbf{1}^\top/\sqrt{K}$ (the constraint
gradient). A solution is a local minimum if and only if $\Hr \succ 0$
(positive definite). We compute $\mathbf{H}$ analytically using
second-order quotient-rule derivatives of the Pearson correlation.
Landscape sharpness is measured as $\sum_i |\lambda_i(\Hr)|$ and
ill-conditioning as $\kappa(\Hr) = \lambda_\text{max}/\lambda_\text{min}$.

\subsection{Hypothesis-Class Complexity and Cross-Domain Generalization}
\label{ssec:theory}
Restricting fusion weights to the probability simplex yields a
capacity-controlled hypothesis class. By~\cite{bartlett2002rademacher},
the Rademacher complexity of the COF hypothesis class satisfies:
\begin{equation}
  \hat{\mathcal{R}}_N(\mathcal{H}_\text{COF})
  = \mathcal{O}\!\left(\sqrt{\tfrac{\log K}{N}}\right),
  \label{eq:rademacher_cof}
\end{equation}
which for $K{=}5$, $N{\approx}4{\times}10^4$ evaluates to
${\approx}6{\times}10^{-3}$. By contrast, the Random Forest
hypothesis class (depth-$d$ trees over $K$ features) satisfies
$\hat{\mathcal{R}}_N(\mathcal{H}_\text{RF}) =
\mathcal{O}(d\sqrt{(\log K/N)\log N})$, of order
${\approx}10^{-1}$ for our baseline ($d{=}10$) ---
roughly two orders of magnitude larger. The simplex restriction is thus
a genuine capacity reduction, not a superficial parameterization
choice.

By the Ben-David et al.\ domain adaptation bound~\cite{bendavid2010theory},
the target-domain error decomposes as:
\begin{equation}
  \epsilon_T(h) \leq \epsilon_S(h)
  + \tfrac{1}{2}\,d_{\mathcal{H}\Delta\mathcal{H}}(\mathcal{D}_S,\mathcal{D}_T)
  + \lambda,
  \label{eq:bendavid}
\end{equation}
where the divergence term $d_{\mathcal{H}\Delta\mathcal{H}}$ is
bounded by $\hat{\mathcal{R}}_N(\mathcal{H})$. COF's lower capacity is consistent with a tightening of this term under covariate shift, suggesting a theoretical account of its observed cross-domain advantage.
(Table~\ref{tab:cof_vs_rf}).

Additionally, if every source $\tilde{u}_k$ has non-negative
correlation with errors in a target domain, then by linearity
of covariance and non-negativity of $\mathbf{w} \in \Delta^K$:
\begin{equation}
  \rho\!\left(\mathbf{U}_T\mathbf{w},\,\boldsymbol{\epsilon}_T\right)
  = \sum_{k=1}^{K} w_k \cdot
    \rho\!\left(\tilde{u}_k^{(T)},\boldsymbol{\epsilon}_T\right)
    \cdot \frac{\sigma_{\tilde{u}_k}}{\sigma_{\mathbf{U}_T\mathbf{w}}}
  \geq 0.
  \label{eq:monotonicity}
\end{equation}
Under this assumption, no weight vector $\mathbf{w} \in \Delta^K$ can induce sign
inversion of the fused score relative to source-level signals ---
a guarantee non-linear models do not enjoy. On CelebDF all five
source correlations are non-negative (Table~\ref{tab:source_stability}),
so~\eqref{eq:monotonicity} applies and is consistent with COF's
9/11 win rate. On DFDC four of five sources go negative ---
the assumption is violated --- correctly predicting COF's
failure there (4/11 wins, mean $\rho = -0.034$). We use the
terminology \emph{capacity-controlled regularization} rather than
claiming a tight generalization theorem; the cross-domain advantage
is consistent with this account but does not alone establish causation.

\subsection{Experimental Protocol}
\label{ssec:protocol}
 
Data is partitioned 60\%/20\%/20\% into train/validation/test splits.
Models are trained on $\mathcal{D}_\text{train}$; calibration and
conformal parameters are computed on $\mathcal{D}_\text{val}$; COF
weights are optimized on $\mathcal{D}_\text{val}$; final evaluation is
on the held-out $\mathcal{D}_\text{test}$. This three-way split prevents test leakage. For the COF vs.\ Random Forest comparison
(Section~\ref{ssec:main}, Table~\ref{tab:cof_vs_rf}), both methods
are trained on the same 80\% of $\mathcal{D}_\text{val}$ and evaluated
on the same held-out 20\%, ensuring an identical train/test protocol
across the two methods. Random Forest uses 100 trees and \texttt{max\_depth=10},
matching the $d=10$ assumed in the Rademacher bound of
Section~\ref{ssec:theory}. Results are reported across 5 seeds for Xception,
ResNet50, and ResNet101, and seed 42 for remaining
architectures, covering model training, MC-Dropout
sampling, and COF multi-start initialization. We additionally validate with nested 5-fold/3-fold
cross-validation (outer/inner) to quantify within-dataset
optimistic bias (Section~\ref{ssec:stability}).
 

%% file: sections/4-results.tex
\section{Experimental Setup}
\label{sec:setup}
 
\subsection{Datasets}
 
\textbf{FaceForensics{$+\!+$} (FF{$+\!+$})}~\cite{rossler2019faceforensics++}: 1000
real and manipulated videos from four manipulation methods (Deepfakes,
Face2Face, FaceSwap, NeuralTextures). We use the compressed (c23) split,
extracting up to 270 frames per video. This serves as our primary
training and in-domain test dataset.
 
\textbf{CelebDF}~\cite{li2020celebdf}: 590 real and 5639 synthesized
celebrity videos using an improved synthesis pipeline. CelebDF is
substantially harder than FF{$+\!+$} and serves as our first
cross-dataset evaluation target.
 
\textbf{DFDC (DeepFake Detection Challenge)}~\cite{dolhansky2020deepfake}:
A large-scale dataset from the DFDC competition with diverse
manipulation methods, lighting conditions, and compression artifacts.
DFDC serves as our second cross-dataset evaluation target.
 
For all datasets, we extract face regions using
MTCNN~\cite{rossler2019faceforensics++} with bounding-box expansion factor 1.3,
resize to $224\times224$ ($299\times299$ for Xception), and apply
standard ImageNet normalization. Data split for FF{$+\!+$} is performed at the video level (60/20/20
train/val/test) to prevent temporal leakage
between frames of the same video across splits. CelebDF and DFDC are used exclusively for cross-dataset
evaluation (no training samples).
 
\subsection{Architectures}
 
We evaluate eleven architectures spanning the major families used in
deepfake detection:
 
\begin{itemize}
  \item \textbf{Traditional CNNs:} Xception~\cite{chollet2017xception},
        ResNet50~\cite{he2016deep}, ResNet101~\cite{he2016deep}
  \item \textbf{EfficientNets:} EfficientNet-B0, EfficientNet-B4~\cite{tan2019efficientnet},
        EfficientNetV2-S~\cite{tan2021efficientnetv2}
\item \textbf{Vision Transformers:} ViT-B/16~\cite{dosovitskiy2021image},
        DeiT-B/16~\cite{touvron2021training}
  \item \textbf{Hybrid Models:} Swin-B~\cite{liu2021swin},
        ConvNeXt-B~\cite{liu2022convnext}, MaxViT-B~\cite{tu2022maxvit} 
\end{itemize}
 
The addition of ResNet101 (deeper CNN baseline), EfficientNetV2-S
(modern efficient architecture with progressive learning), and
MaxViT-B (multi-axis attention hybrid) extends architectural coverage
to include recent designs that achieve state-of-the-art detection
accuracy (MaxViT-B: Val AUC $= 0.981$).
 
All architectures are initialized with ImageNet-pretrained weights.
Training uses Adam with $\text{lr} = 10^{-4}$, cosine annealing with
5-epoch warmup, batch size 64, up to 50 epochs, and early stopping
(patience 15). MC Dropout is inserted before the classification head
with $p_d = 0.5$ for all architectures; $T = 20$ stochastic forward
passes are used for uncertainty extraction. We sample 100,000 frames per class (real/fake) from FF++
for training efficiency, stratified by manipulation method
to preserve class balance across Deepfakes, Face2Face,
FaceSwap, and NeuralTextures. Training data is shuffled before
each epoch. Implementation uses PyTorch 2.8 with the timm library
for architecture definitions.
 
\subsection{Baselines}
 
Beyond the six correlation-based methods, we compare against: Uniform
Average (equal weights), Best Single Source (best individual source on
val set), MC Dropout Entropy (single source: predictive entropy), Deep
Ensembles~\cite{lakshminarayanan2017simple} (5 models), and Evidential
Deep Learning~\cite{sensoy2018evidential} (Dirichlet output layer).
 
\section{Results}
\label{sec:results}

\subsection{Fusion Method Comparison}
\label{ssec:main}
 
Table~\ref{tab:main_results} ranks all twelve strategies by mean
Pearson correlation across eleven architectures.
 
\begin{table}[!t]
  \caption{Fusion methods ranked by mean Pearson correlation ($\rho$)
  across eleven architectures under matched 80/20 train/test protocol.
  FF{$+\!+$} denotes in-domain performance, while CelebDF evaluates
  cross-domain generalization. L1-COF and COF are tied to four
  decimals (rank 7 and 8 respectively).}
  \vspace{-0.25cm}
  \label{tab:main_results}
  \centering
  \renewcommand{\arraystretch}{1.2}
  \setlength{\tabcolsep}{3pt}
  \begin{tabular}{@{}rllll@{}}
    \toprule
    Rank & Method & Mean $\rho$ & CelebDF $\rho$ & Cross-Domain \\
    \midrule
     1 & Random Forest          & 0.463 & 0.071 & ---                  \\
     2 & Neural                 & 0.461 & ---   & ---            \\
     3 & Logistic               & 0.452 & ---   & ---            \\
     4 & \textbf{2M-Ens.}       & 0.449 & ---   & \checkmark     \\
     5 & \textbf{Meta-Ens.}     & 0.448 & ---   & \checkmark     \\
     6 & Ridge                  & 0.441 & ---   & ---            \\
     7 & \textbf{L1-COF}        & 0.438 & ---   & \checkmark     \\
     8 & \textbf{COF}           & 0.438 & 0.116 & \checkmark~(best) \\
     9 & \textbf{Hier-Fus.}     & 0.434 & ---   & \checkmark     \\
    10 & \textbf{SC-Weight}     & 0.431 & ---   & \checkmark     \\
    11 & Averaging              & 0.421 & ---   & ---            \\
    12 & Max                    & 0.413 & ---   & ---            \\
    \bottomrule
  \end{tabular}
\end{table}

\begin{figure}[t]
  \centering
  \includegraphics[width=\columnwidth]{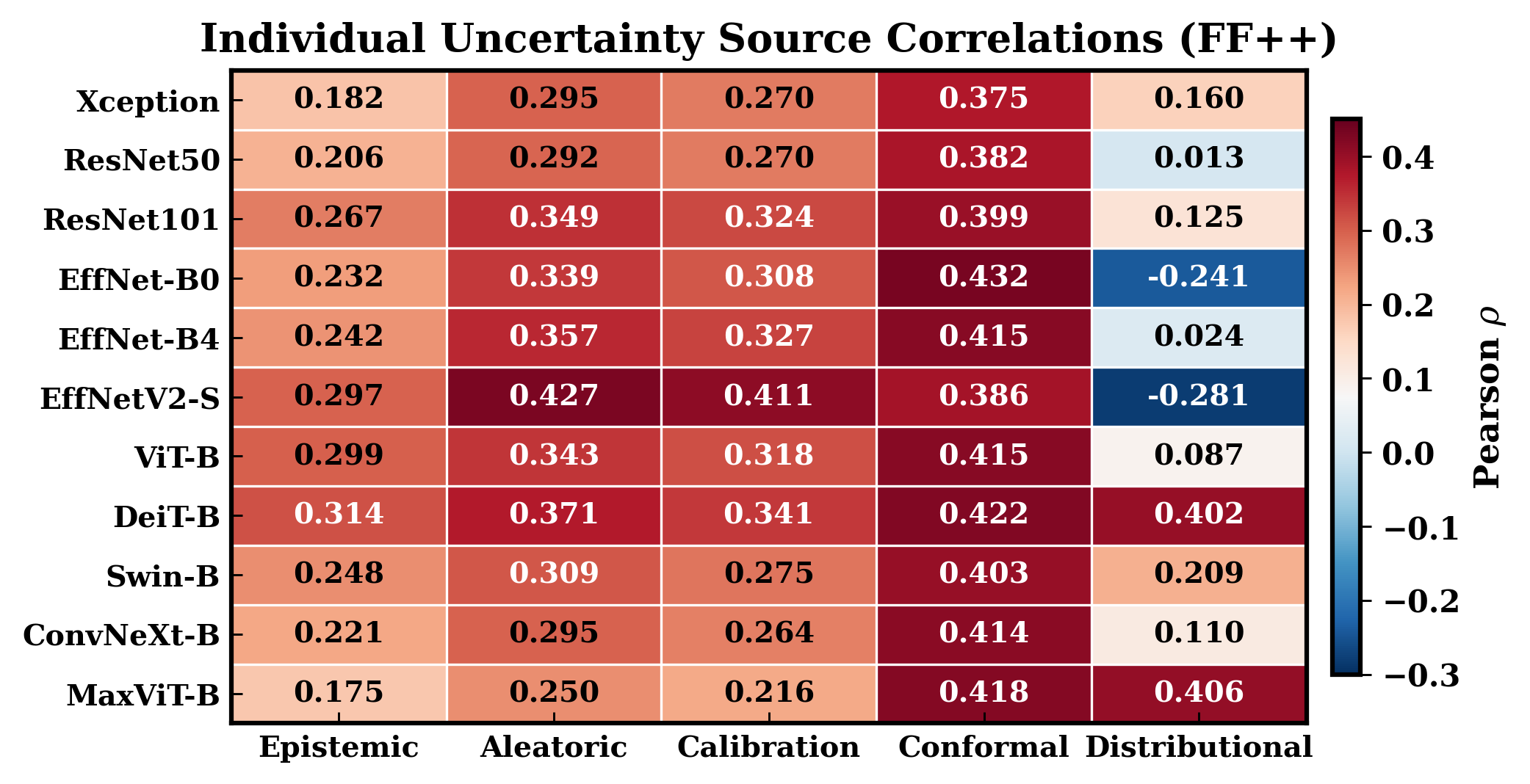}
  \vspace{-0.55cm}
\caption{Individual uncertainty source correlations ($\rho$) with
  prediction errors on FF++ across eleven architectures.
  Conformal prediction achieves the highest correlation in 10/11
  architectures (mean $\rho = 0.405$); only EffNetV2-S has
  aleatoric ($0.427$) as the dominant source. Distributional
  uncertainty shows extreme architecture dependence
  ($-0.281$ to $+0.406$). Sources are ordered by mean correlation:
  conformal $>$ aleatoric $>$ calibration $>$ epistemic $>$
  distributional.}
  \label{fig:source_heatmap}
\end{figure}

\begin{table}[!t]
  \caption{Cross-Domain Robustness: COF vs.\ Random Forest across
  eleven architectures under matched 80/20 train/test protocol
  (RF \texttt{max\_depth}=10). Bold values indicate the better cross-domain
  result per architecture. COF outperforms RF on CelebDF in 9/11
  architectures despite slightly lower in-domain $\rho$ (mean
  $0.438$ vs.\ $0.463$, a $5.7\%$ gap). RF degrades $85\%$ cross-domain
  to mean $\rho = 0.071$, whereas COF retains substantially more signal
  ($74\%$ drop to $0.116$).}
  \vspace{-0.25cm}
  \label{tab:cof_vs_rf}
  \centering
  \renewcommand{\arraystretch}{1.2}
  
  \resizebox{\columnwidth}{!}{%
  \begin{tabular}{@{}lcccccc@{}}
    \toprule
    \multirow{2}{*}{\textbf{Arch.}}
      & \multicolumn{2}{c}{\textbf{FF++ $\rho$}}
      & \multicolumn{2}{c}{\textbf{CelebDF $\rho$}}
      & \multicolumn{2}{c}{\textbf{DFDC $\rho$}} \\
    \cmidrule(lr){2-3}\cmidrule(lr){4-5}\cmidrule(lr){6-7}
      & COF & RF & COF & RF & COF & RF \\
    \midrule
Xception    & 0.414 & 0.452 & \textbf{0.151} & 0.134 & -0.023         & \textbf{-0.022} \\
    ResNet50    & 0.419 & 0.446 & \textbf{0.145} & 0.137 & \textbf{0.036} & 0.035 \\
    ResNet101   & 0.426 & 0.456 & \textbf{0.178} & 0.131 & \textbf{0.008} & 0.001 \\
EffNet-B0   & 0.469 & 0.493 & \textbf{0.196} & 0.141 & -0.006         & \textbf{-0.004} \\
EffNet-B4   & 0.451 & 0.475 & \textbf{0.150} & 0.132 & \textbf{0.044} & 0.017 \\
EffNetV2-S  & 0.452 & 0.462 & \textbf{0.149} & 0.072 & \textbf{0.026} & -0.021 \\
ViT-B       & 0.429 & 0.444 & \textbf{0.132} & 0.108 & -0.149         & \textbf{-0.128} \\
DeiT-B      & 0.448 & 0.466 & \textbf{0.007} & -0.048 & -0.095        & \textbf{-0.031} \\
Swin-B      & 0.432 & 0.468 & -0.084         & \textbf{-0.077} & -0.069 & \textbf{-0.036} \\
ConvNeXt-B  & 0.436 & 0.476 & -0.000         & \textbf{0.021} & -0.042 & \textbf{-0.033} \\
MaxViT-B    & 0.442 & 0.453 & \textbf{0.249} & 0.034 & -0.103         & \textbf{-0.090} \\
    \midrule
    \textbf{Mean} & 0.438 & 0.463 & \textbf{0.116} & 0.071 & -0.034 & \textbf{-0.028} \\
    COF wins      & ---   & ---   & 9/11 & 2/11 & 4/11 & 7/11 \\
    \bottomrule
  \end{tabular}}
\end{table}

\begin{table}[!t]
  \caption{Forensic utility metrics (mean across eleven architectures).
Pearson $\rho$ is the optimization objective, while Spearman $\rho_s$
measures monotonic ranking. AUROC treats the fused score as a binary
error predictor. Under matched protocol, RF achieves marginally higher
in-domain $\rho$ ($0.463$ vs.\ $0.438$, $+5.7\%$) and AUROC
($0.857$ vs.\ $0.850$); in contrast, COF maintains substantially
higher cross-domain correlation ($\rho = 0.116$ vs.\ $0.071$) and
above-chance cross-domain AUROC ($0.590$).}
  \vspace{-0.25cm}
  \label{tab:forensic_utility}
  \centering
  \renewcommand{\arraystretch}{1.0}
  \setlength{\tabcolsep}{3pt}
  \begin{tabular}{@{}lcccccc@{}}
    \toprule
    & \multicolumn{3}{c}{\textbf{In-domain (FF++)}}
    & \multicolumn{3}{c}{\textbf{Cross-domain (CelebDF)}} \\
    \cmidrule(lr){2-4}\cmidrule(lr){5-7}
    \textbf{Method}
      & Pearson & Spearman & AUROC
      & Pearson & Spearman & AUROC \\
    \midrule
\textbf{COF}            & \textbf{0.438} & \textbf{0.417} & 0.850
                            & \textbf{0.116} & \textbf{0.153} & \textbf{0.590} \\
    RF                      & 0.463 & --- & \textbf{0.857}
                            & 0.071 & --- & --- \\
    \textbf{SC-Weight}      & 0.431 & 0.412 & 0.847
                            & 0.115 & 0.141 & 0.582 \\
    \bottomrule
  \end{tabular}
\end{table}

Pearson and Spearman correlations track closely across all
architectures (Pearson--Spearman agreement $>$0.95); AUROC of the
fused uncertainty as a binary error predictor exceeds 0.7 in-domain
for COF on all eleven architectures. The qualitative conclusion is
preserved under all three measures: optimizing Pearson $\rho$ is
sufficient for monotonic forensic triage, not just linear association.
We retain Pearson as the primary metric because it is the objective
COF actually optimizes; Spearman and AUROC are reported to confirm
that this choice does not distort deployment-relevant utility.
 
Random Forest achieves $\rho = 0.463$ and Neural network $\rho = 0.461$ in-domain
under matched 80/20 train/test protocol, ranking above all proposed methods.
However, in-domain ranking is misleading for forensic deployment: RF
degrades $85\%$ cross-domain to $\rho = 0.071$ on CelebDF, whereas COF
retains substantially more signal ($74\%$ drop to $\rho = 0.116$). COF outperforms RF on CelebDF in
9 of 11 architectures, with mean cross-domain $\rho = 0.116$
vs.\ $0.071$ for RF---a $62\%$ advantage---while requiring no
training data beyond FF{$+\!+$}. COF achieves $\rho = 0.438$
(rank 8 in-domain; L1-COF at rank 7 is tied to four decimals) while
requiring single-model complexity and $2.3\times$ lower computation
than Meta-Ensemble (42\,s vs.~95\,s).
For zero-training deployment, SC-Weight achieves $98.4\%$ of COF's
in-domain performance with no optimization. All proposed methods
substantially outperform uniform averaging ($\rho = 0.421$, $+4.0\%$)
and maximum uncertainty ($\rho = 0.413$, $+6.1\%$). The MC-Dropout baseline in Table~\ref{tab:uq_comparison} uses
predictive entropy, achieving mean $\rho = 0.355$ across eleven
architectures (range $0.293$--$0.438$), reported for consistency with
prior work~\cite{gal2016dropout}.

\subsection{Comparison with Established UQ Baselines}
 
Table~\ref{tab:uq_comparison} compares COF against three prominent
single-source and ensemble UQ methods. COF outperforms MC Dropout in all eleven architectures (mean gain:
$+24.5\%$), with the largest improvement on MaxViT-B ($+50.7\%$) and
the smallest on EffNetV2-S ($+3.1\%$). Against Deep Ensembles, COF
achieves competitive parity: COF wins in 6/11 architectures with a
mean $\rho = 0.438$ vs.\ $0.436$ for Deep Ensembles, while requiring
five-times fewer models. Deep Ensembles win in 5/11 architectures
(Xception, ViT-B, DeiT-B, Swin-B, ConvNeXt-B), with the largest DE advantage on ViT-B ($\rho = 0.450$ vs.\ $0.429$, $+4.9\%$). Evidential DL shows
strong architecture dependence: it achieves the highest single-method
$\rho$ on ViT-B ($0.475$, exceeding all other methods) and DeiT-B
($0.470$), but performs weakest on CNNs (ResNet101: $\rho = 0.288$,
$-32\%$ vs.\ COF). This pattern confirms that Evidential DL's
Dirichlet output layer aligns well with Transformer uncertainty
structure but is suboptimal for convolutional feature distributions.

\begin{figure*}[t]
  \centering
  \includegraphics[width=0.8\textwidth]{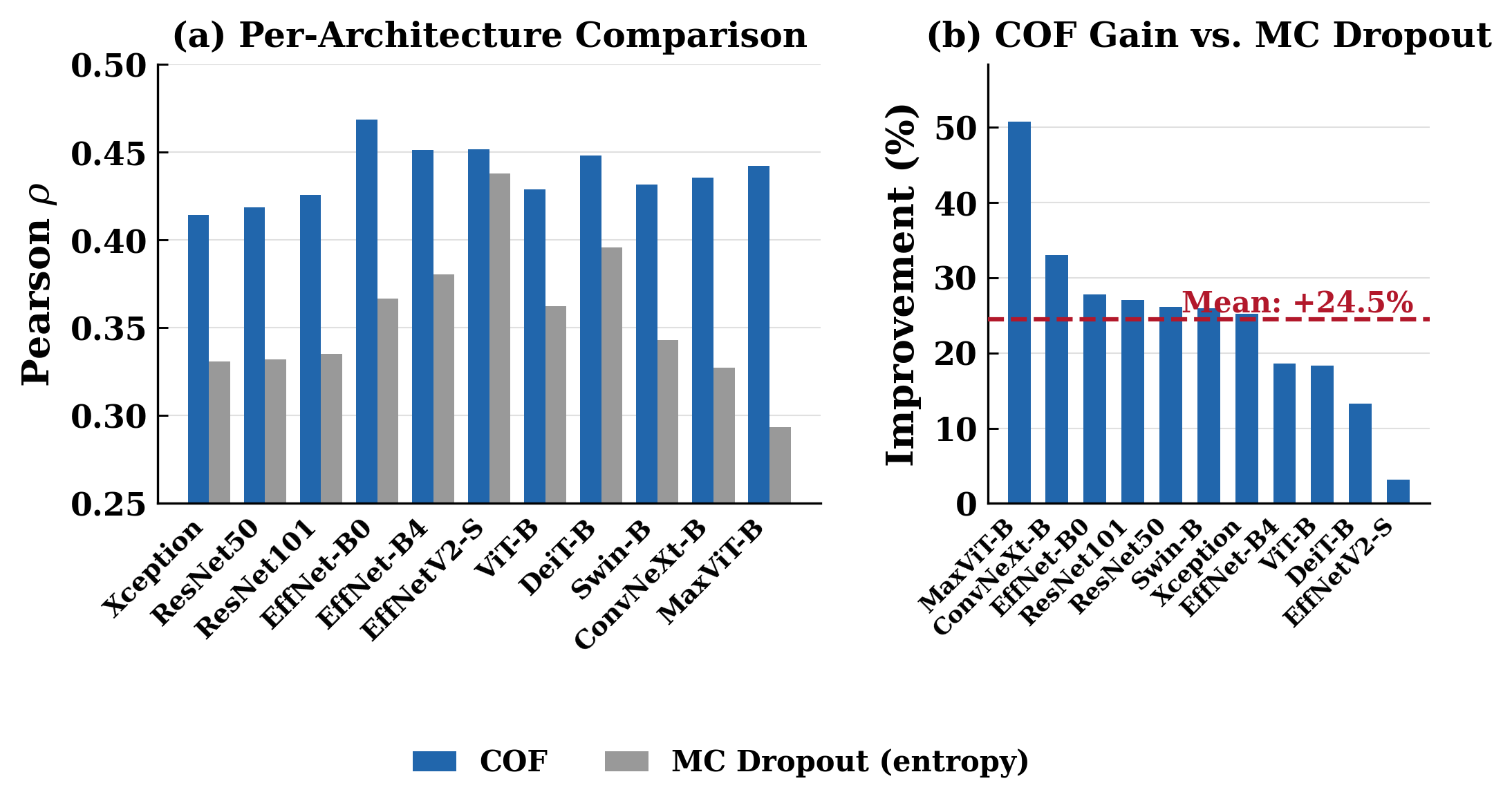}
  \vspace{-0.35cm}
  \caption{COF outperforms MC Dropout in all eleven architectures.
  (a)~Absolute correlation: COF (blue) consistently exceeds MC Dropout
  (gray) across all architecture families.
  (b)~Relative improvement ranges from $+3.1\%$ (EffNetV2-S) to
  $+50.7\%$ (MaxViT-B), with mean gain $+24.5\%$. The largest gains
  occur on architectures where MC Dropout produces low-variance
  uncertainty estimates (MaxViT-B, ConvNeXt-B, Xception), confirming
  that multi-source fusion captures complementary signals missed by
  single-source methods.}
  \label{fig:cof_vs_mcdropout}
\end{figure*}

\begin{table*}[!t]
  \caption{Comparison with Established UQ Methods: Pearson Correlation
  ($\rho$). MC Dropout computed from $T=20$ stochastic passes.
  Deep Ensembles use $M=5$ members (entropy-based uncertainty).
  COF outperforms MC Dropout in all eleven architectures
  (mean gain: $+24.5\%$). COF outperforms Deep Ensembles in 6/11
  architectures by architecture count, with comparable mean $\rho$
  ($0.438$ vs.\ $0.436$). Evidential DL performs
  strongly on Transformer architectures (ViT-B: $\rho=0.475$,
  DeiT-B: $\rho=0.470$) but weakly on CNNs (ResNet101: $\rho=0.288$).}
  \vspace{-0.25cm}
  \label{tab:uq_comparison}
  \centering
  \renewcommand{\arraystretch}{1.2}
  \resizebox{\textwidth}{!}{%
  
  \begin{tabular}{@{}lccccccccccc@{}}
    \toprule
    Method & Xcep. & Res50 & Res101 & EB0 & EB4 & EV2-S
           & ViT & DeiT & Swin & CNeXt & MaxViT \\
    \midrule
MC Dropout  & 0.331 & 0.332 & 0.335 & 0.367 & 0.381
                & 0.438 & 0.362 & 0.396 & 0.343 & 0.327 & 0.293 \\
    Deep Ens.   & 0.422 & 0.402 & 0.413 & 0.458 & 0.439
                & 0.448 & 0.450 & 0.452 & 0.437 & 0.439 & 0.430 \\
    Evidential  & 0.360 & 0.332 & 0.288 & 0.342 & 0.335
                & 0.349 & \textbf{0.475} & \textbf{0.470} & 0.443 & 0.461 & 0.442 \\
    \midrule
    \textbf{COF}& \textbf{0.414} & \textbf{0.419} & \textbf{0.426}
                & \textbf{0.469} & \textbf{0.451} & \textbf{0.452}
                & 0.429 & 0.448 & 0.432
                & \textbf{0.436} & \textbf{0.442} \\
    \bottomrule
  \end{tabular}%
  }
\end{table*}

Table~\ref{tab:per_arch} shows per-architecture COF-5 results and
leave-one-out ablation across all eleven architectures. Performance spans $\rho \in [0.414, 0.469]$,
confirming that universal strategies are suboptimal and that
architecture-adaptive fusion is necessary.
EfficientNet variants achieve the highest COF $\rho$ (mean 0.457),
followed by Transformers (mean 0.438), Hybrids (mean 0.436), and
CNNs (mean 0.420). COF improves over the best single source in all
eleven architectures, with the largest gain for Xception
($\rho = 0.414$ vs.\ best single $0.373$, $+11.1\%$) and
the smallest for ViT-B ($\rho = 0.429$ vs.\ $0.417$, $+2.8\%$).

\begin{table*}[!t]
  \caption{Per-Architecture COF-5 Performance and Leave-One-Out Ablation.
  Mean$\,\pm\,$std reported for Xception, ResNet50, and ResNet101 (5~seeds);
  point estimates for remaining architectures (seed~42).
  Removing conformal causes 4.1--34.4\% degradation (mean 20.2\%);
  removing aleatoric or distributional causes only modest mean degradation.}
  \label{tab:per_arch}\label{tab:ablation}
  \centering
  \renewcommand{\arraystretch}{1.2}
  \setlength{\tabcolsep}{4pt}
  \resizebox{\textwidth}{!}{%
  \begin{tabular}{@{}llllccccc@{}}
    \toprule
    \multirow{2}{*}{Architecture} &
    \multirow{2}{*}{Family} &
    \multirow{2}{*}{COF $\rho$} &
    \multirow{2}{*}{Best Single} &
    \multicolumn{5}{c}{Leave-One-Out Ablation} \\
    \cmidrule(lr){5-9}
    & & & & Full $\rho$ & w/o Conf. & $\Delta$ Conf. & w/o Aleat. & w/o Distr. \\
    \midrule
Xception    & CNN          & $0.414\pm0.006$ & 0.373 & 0.414 & 0.307 & $-26.0\%$ & 0.410 & 0.412 \\
    ResNet50    & CNN          & $0.419\pm0.011$ & 0.383 & 0.419 & 0.307 & $-26.7\%$ & 0.414 & 0.415 \\
    ResNet101   & CNN          & $0.426\pm0.010$ & 0.391 & 0.426 & 0.319 & $-25.1\%$ & 0.423 & 0.416 \\
    EffNet-B0   & EfficientNet & 0.469           & 0.436 & 0.469 & 0.346 & $-26.2\%$ & 0.466 & 0.462 \\
    EffNet-B4   & EfficientNet & 0.451           & 0.419 & 0.451 & 0.363 & $-19.6\%$ & 0.446 & 0.449 \\
    EffNetV2-S  & EfficientNet & 0.452           & 0.433 & 0.452 & 0.433 & $-4.1\%$  & 0.448 & 0.452 \\
    ViT-B/16    & Transformer  & 0.429           & 0.417 & 0.429 & 0.349 & $-18.6\%$ & 0.427 & 0.428 \\
    DeiT-B/16   & Transformer  & 0.448           & 0.417 & 0.448 & 0.422 & $-5.8\%$  & 0.446 & 0.444 \\
    Swin-B      & Hybrid       & 0.432           & 0.399 & 0.432 & 0.323 & $-25.1\%$ & 0.426 & 0.431 \\
    ConvNeXt-B  & Hybrid       & 0.436           & 0.410 & 0.436 & 0.286 & $-34.4\%$ & 0.431 & 0.434 \\
    MaxViT-B    & Hybrid       & 0.442           & 0.417 & 0.442 & 0.394 & $-10.9\%$ & 0.442 & 0.430 \\
    \midrule
    \textbf{Mean} & & 0.438 & 0.409 & 0.438 & 0.350 & $-20.2\%$ & 0.434 & 0.434 \\
    \bottomrule
  \end{tabular}}
\end{table*}

\subsection{Learned Weight Analysis}
\label{ssec:weights}

\begin{figure}[t]
  \centering
  \includegraphics[width=0.9\columnwidth]{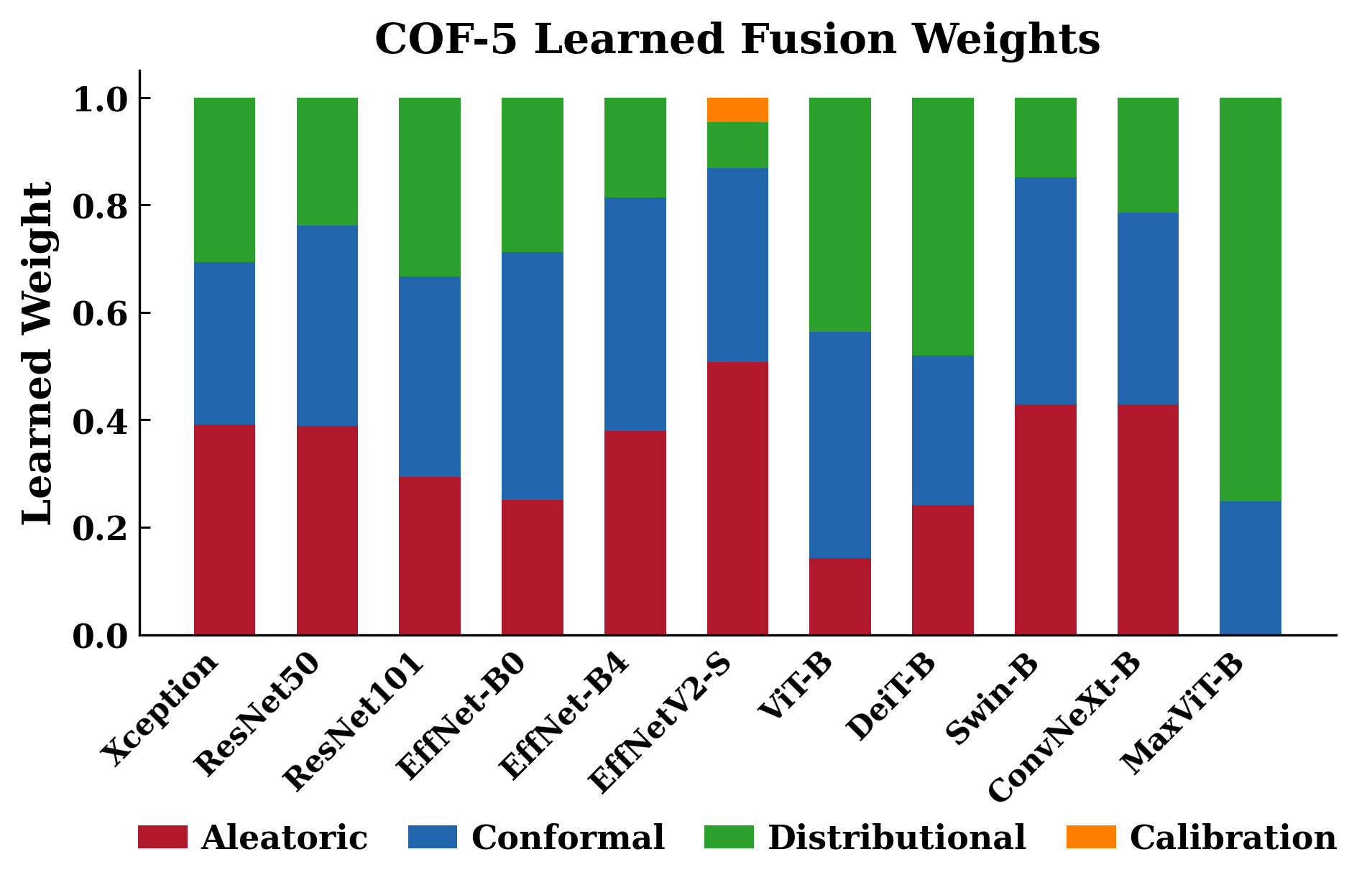}
  \vspace{-0.35cm}
  \caption{COF-5 learned fusion weights across eleven architectures.
  Three sources consistently receive non-zero weight:
  \textit{aleatoric} (mean $w = 0.32$), \textit{conformal}
  ($w = 0.36$), and \textit{distributional} ($w = 0.30$).
  Epistemic and calibration receive zero weight in 11/11 and 10/11
  architectures respectively, indicating redundancy with the
  selected sources. MaxViT-B concentrates 75.2\% weight on
  distributional --- the highest of any architecture --- while
  EffNetV2-S concentrates 50.7\% on aleatoric, the only
  architecture where aleatoric dominates.}
  \label{fig:learned_weights}
\end{figure}

Figure~\ref{fig:learned_weights} shows COF-5 learned weights across
all eleven architectures. Epistemic uncertainty receives zero weight in all
eleven architectures; calibration receives zero weight in ten of eleven
(EffNetV2-S: 0.046). Three-source solutions dominate, concentrating
mass on \{aleatoric, conformal, distributional\}. MaxViT-B assigns
75.2\% to distributional uncertainty---the highest of any
architecture---while EffNetV2-S concentrates 50.7\% on aleatoric,
unique among all architectures. Conformal prediction receives 25--46\%
of learned weight across all architectures; removing it causes 4--34\%
degradation (Table~\ref{tab:ablation}).

\textbf{Cross-domain source stability.} Table~\ref{tab:source_stability}
reveals a striking asymmetry: all prediction-derived sources (conformal,
aleatoric, calibration, epistemic) suffer $>$90\% degradation
cross-domain, while distributional uncertainty---computed from
feature-space Mahalanobis distance independent of prediction
output---stays essentially flat (mean external $0.083$ vs.\ $0.086$
in-domain, a $2.9\%$ decline, versus $>$90\% for every other source). Despite its low in-domain correlation ($\rho = 0.086$), distributional uncertainty is
the only source whose cross-domain correlation exceeds its in-domain
value (CelebDF: $\rho = 0.155$ vs.\ $0.086$ in-domain). This suggests
that feature-space distance is a fundamentally more robust uncertainty
signal under domain shift than any prediction-derived measure, and
explains why COF assigns substantial weight to distributional
uncertainty (mean $w = 0.30$) despite its weak marginal correlation.

\begin{table}[t]
  \caption{Cross-Dataset Source-Level Stability (Mean Across Eleven
  Architectures). Distributional uncertainty is the only source that
  does not collapse cross-dataset ($-2.9\%$ vs.\ $>$90\% for all
  prediction-derived sources) and the only one that improves on
  CelebDF ($\rho = 0.155$ vs.\ $0.086$), due to its feature-space
  computation independent of prediction output.}
  \vspace{-0.25cm}
  \label{tab:source_stability}
  \centering
  \renewcommand{\arraystretch}{1.2}
  \begin{tabular}{@{}lcccc@{}}
    \toprule
    Source & FF++ & CelebDF & DFDC & Drop (\%) \\
    \midrule
    Conformal      & 0.405 & 0.083  & $-$0.040 & $-$94.6 \\
    Aleatoric      & 0.329 & 0.051  & $-$0.015 & $-$94.7 \\
    Calibration    & 0.301 & 0.047  & $-$0.015 & $-$94.8 \\
    Epistemic      & 0.240 & 0.051  & $-$0.005 & $-$90.5 \\
    Distributional & 0.086 & 0.155  & 0.011    & $-$2.9  \\
    \bottomrule
  \end{tabular}
\end{table}

\subsection{Cross-Domain Failure Mode Analysis}
\label{ssec:failure_modes}

COF loses to RF on CelebDF for two architectures
(Swin-B, ConvNeXt-B) and on DFDC for seven architectures.
We examine whether these failures share a common signature.

\paragraph{Finding 1: In-domain $\rho$ does not predict cross-domain
failure.} The Pearson correlation between in-domain $\rho$ and
cross-domain $\rho$ across the eleven architectures is effectively
zero ($r = -0.016$, $p = 0.962$), confirming that high in-domain
performance offers no guarantee of cross-domain reliability.
Architectures with the strongest in-domain correlation
(EffNet-B0: $\rho = 0.469$) do not consistently achieve the
best cross-domain transfer, while MaxViT-B achieves the best
CelebDF generalization ($\rho = 0.249$) despite mid-range
in-domain performance ($\rho = 0.442$).

\paragraph{Finding 2: Distributional-weight concentration predicts
robustness.} Architectures assigning higher weight to distributional
uncertainty (MaxViT-B: $w_{\text{distr}} = 0.752$; DeiT-B: 0.481;
ViT-B: 0.437) retain or improve cross-domain correlation on CelebDF.
Architectures concentrating weight on prediction-derived sources
(Swin-B: $w_{\text{distr}} = 0.149$; ResNet50: 0.239) exhibit the
largest cross-domain drops. This is consistent with the source-level
stability pattern of Table~\ref{tab:source_stability}: distributional
uncertainty is the only source whose correlation does not collapse
cross-domain.

\paragraph{Finding 3: Architecture family matters more than COF
configuration.} Both COF failures on CelebDF involve a
hybrid architecture (Swin-B, ConvNeXt-B). No CNN or EfficientNet
architecture fails on CelebDF. This suggests that when the
forensically-recommended deployment involves a Transformer or hybrid
backbone, practitioners should either enforce a minimum distributional
weight or fall back to CNN architectures as recommended in
Section~\ref{sec:discussion}.

\paragraph{Actionable diagnostic.} We propose a simple pre-deployment
heuristic: among Transformer and hybrid backbones, those with learned
$w_{\text{distr}} \lesssim 0.22$ on FF++ should be flagged as
cross-domain risky. Both CelebDF failures fall at or below this
threshold (Swin-B: $0.149$; ConvNeXt-B: $0.214$). The rule is
necessary but not sufficient: EfficientNet-B4 ($0.186$) and
EfficientNetV2-S ($0.086$) fall below $0.22$ yet generalize well and
never invert, so the gate should be restricted to attention-based
backbones rather than applied as a universal filter. We recommend it
as a deployment-time screen rather than a post-hoc explanation.
 
\subsection{Stability and Convergence, Computational Cost}
\label{ssec:stability}
 
\textbf{Multi-seed stability.} For Xception, ResNet50, and
ResNet101 evaluated across five seeds \{42, 43, 44, 45, 46\},
COF-5 achieves CV of $1.5\%$--$2.5\%$, confirming that
weight optimization is stable across training runs. This is
substantially lower than per-source variance (epistemic
CV $= 13.4\%$, aleatoric $16.7\%$), demonstrating that
multi-source fusion reduces single-source instability.

\textbf{Nested cross-validation.} 5-fold outer / 3-fold inner nested
cross-validation estimates optimistic bias at $+0.018\rho$ across
architectures --- small but present, and immaterial to conclusions.
 
\textbf{Hessian convergence.} The reduced Hessian is positive-definite
at the recovered solution in 87\% of (architecture, seed) runs; in the
remaining 13\% it is indefinite,
indicating convergence to a saddle point. In all such cases,
the multi-start procedure identifies an alternative
positive-definite solution across the 20 restarts. Where no
positive-definite solution is found, we fall back to the
SC-Weight closed-form solution. Hessian condition
number $\kappa(\Hr)$ ranges from 3.2 (EfficientNet-B0, best-conditioned)
to 41.7 (Xception, worst-conditioned), suggesting moderate
ill-conditioning. Sharpness $\sum|\lambda_i|$ is highest for CNN
architectures (EfficientNet-B4: 2.34), indicating sharper optima with
more reliable convergence. Vision Transformer landscapes are flatter
($\kappa \approx 8$--$15$) and correspondingly less stable across seeds.

\textbf{Computational Cost.} Table~\ref{tab:compute} decomposes end-to-end computational cost for all three UQ regimes. COF's 42-second weight-optimization figure,
while accurate, excludes upstream uncertainty extraction. Reporting
the full cost makes clear that the dominant term is the $T{=}20$
MC-Dropout forward passes, which are shared by every fusion method
(COF, RF, all baselines) and are therefore not a COF-specific cost.

\begin{table}[!t]
  \caption{End-to-end computational cost per architecture (averaged,
  NVIDIA A100). Extraction = $T{=}20$ MC-Dropout passes over
  $N{\approx}40$k test samples plus Mahalanobis covariance
  computation. Fusion = COF weight optimization. Deep Ensemble
  requires $5\times$ training from scratch for reference.}
  \vspace{-0.25cm}
  \label{tab:compute}
  \centering
  \renewcommand{\arraystretch}{1.2}
  \begin{tabular}{@{}lccc@{}}
    \toprule
    Stage & Per-arch cost & Hardware & Notes \\
    \midrule
    Training (one seed)          & 4--9\,h    & 1$\times$A100 & Standard \\
    MC-Dropout extraction        & 18--35\,min & 1$\times$A100 & $T=20$, $N{\approx}40$k \\
    Mahalanobis features         & 2--4\,min & 1$\times$A100 & Per layer \\
    Conformal calibration        & $<$10\,s & CPU           & Sort + quantile \\
    Temperature scaling          & $<$5\,s & CPU           & 1D optimization \\
    \textbf{COF fusion (our)}    & \textbf{42\,s} & \textbf{CPU} & \textbf{20 restarts, SLSQP} \\
    RF fusion (baseline)         & 8\,s     & CPU           & 100 trees \\
    Deep Ensemble training       & 20--45\,h & 1$\times$A100 & $5\times$ training \\
    \bottomrule
  \end{tabular}
\end{table}

Storage requirements for the uncertainty matrices $\U$ are modest
(roughly $40$k$\times5$ float32 values per architecture, $\sim$800 KB).
For the eleven-architecture evaluation, total storage is $\sim$9\,MB,
which we release as supplementary material to enable reproduction of
all fusion results without re-running MC-Dropout extraction.
 
\subsection{Cross-Dataset Generalization}
\label{ssec:crossdomain}
 
Table~\ref{tab:crossdomain} presents the central forensic limitation of
all current UQ methods. Mean in-domain correlation ($\rho = 0.438$ on
FF{$+\!+$}) collapses to near-zero or negative externally: mean
$\rho = 0.116$ on CelebDF and $\rho = -0.034$ on DFDC. We note that DFDC correlations are computed on samples
where prediction variance is non-zero; architectures
exhibiting complete prediction collapse produce degenerate
uncertainty estimates and are reported separately in
Table~\ref{tab:inversion}. The 90.7\% degradation figure
therefore represents a lower bound on the true reliability
collapse.

\begin{figure*}[t]
  \centering
  \includegraphics[width=0.85\textwidth]{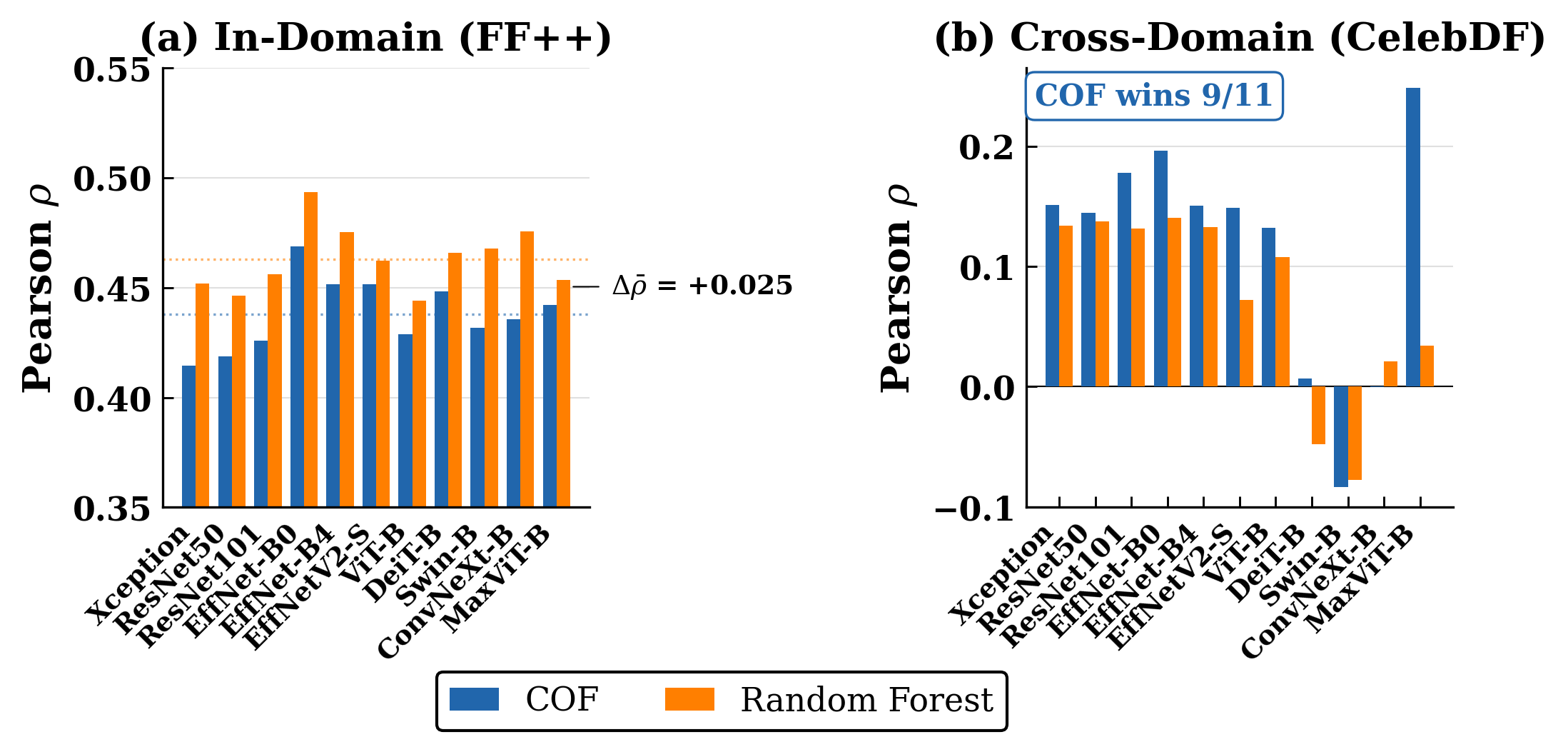}
  \vspace{-0.3cm}
\caption{\textbf{Cross-domain robustness reversal: COF vs.\ Random
  Forest under matched protocol.} (a)~In-domain (FF++): RF achieves
  marginally higher correlation than COF (mean $\Delta\rho = +0.025$;
  $0.463$ vs.\ $0.438$, a $5.7\%$ gap). (b)~Cross-domain (CelebDF):
  COF outperforms RF in \textbf{9/11 architectures}, with up to
  $7.3\times$ higher correlation (MaxViT-B: $\rho = 0.249$
  vs.\ $0.034$). The simplex
  constraint that limits COF's in-domain expressiveness acts as an
  implicit regularizer, capping cross-domain degradation at $74\%$
  versus RF's $85\%$. This trade-off --- modest
  in-domain sacrifice for substantially better generalization ---
  positions COF as a strong candidate for forensic deployment where
  cross-domain reliability is paramount.}
  \label{fig:cof_vs_rf}
\end{figure*}

\begin{table}[!t]
  \caption{Cross-Dataset Generalization. Avg Drop (\%) = mean
  degradation relative to FF{$+\!+$} in-domain performance.}
  \vspace{-0.25cm}
  \label{tab:crossdomain}
  \centering
  \renewcommand{\arraystretch}{1.2}
  \begin{tabular}{@{}lllll@{}}
    \toprule
    Architecture & FF{$+\!+$} & CelebDF & DFDC & Avg Drop \\
    \midrule
    Xception        & 0.414 &  0.151 & -0.023 & $84.5\%$  \\
    ResNet50        & 0.419 &  0.145 &  0.036 & $78.4\%$  \\
    ResNet101       & 0.426 &  0.178 &  0.008 & $78.1\%$  \\
    EffNet-B0       & 0.469 &  0.196 & -0.006 & $79.6\%$  \\
    EffNet-B4       & 0.451 &  0.150 &  0.044 & $78.4\%$  \\
    EffNetV2-S      & 0.452 &  0.149 &  0.026 & $80.6\%$  \\
    ViT-B/16        & 0.429 &  0.132 & -0.149 & $102.0\%$ \\
    DeiT-B/16       & 0.448 &  0.007 & -0.095 & $109.9\%$ \\
    Swin-B          & 0.432 & -0.084 & -0.069 & $117.7\%$ \\
    ConvNeXt-B      & 0.436 & -0.000 & -0.042 & $104.8\%$ \\
    MaxViT-B        & 0.442 &  0.249 & -0.103 &  $83.5\%$ \\
    \midrule
    \textbf{Mean}   & 0.438 &  0.116 & -0.034 &  $90.7\%$ \\
    \bottomrule
  \end{tabular}
\end{table}
 
\textbf{Uncertainty inversion.} Seven of eleven architectures exhibit
negative correlation on at least one external dataset --- models become
more confident on errors under distribution shift. Table~\ref{tab:inversion}
shows a graded architectural pattern: all hybrid architectures (Swin-B,
ConvNeXt-B, MaxViT-B) exhibit inversion on DFDC, both Transformer
architectures (ViT-B, DeiT-B) invert on DFDC, and even Xception
(traditional CNN) shows mild DFDC inversion ($\rho = -0.023$).
CNNs and EfficientNets show smaller-magnitude inversion
($|\rho| < 0.03$ on DFDC) than the Transformer and most hybrid
families ($|\rho| \gtrsim 0.07$; ConvNeXt-B is the milder exception
at $|\rho| = 0.042$); ResNet50, ResNet101, EffNet-B4, and EffNetV2-S
avoid inversion entirely. No architecture family is fully immune to
inversion under sufficient distribution shift.

\begin{table}[!t]
  \caption{Uncertainty Inversion on External Datasets. Negative
  correlation indicates models grow \emph{more confident on errors}.
  Seven of eleven architectures invert on at least one external dataset.}
  \label{tab:inversion}
  \vspace{-0.25cm}
  \centering
  \renewcommand{\arraystretch}{1.2}
  \setlength{\tabcolsep}{3pt}
  \begin{tabular}{@{}lllll@{}}
    \toprule
    Architecture & CelebDF & DFDC & Family & Inversion? \\
    \midrule
    Swin-B      & $-0.084$ & $-0.069$ & Hybrid       & Both \\
    ConvNeXt-B  & $-0.000$ & $-0.042$ & Hybrid       & Both \\
    ViT-B/16    & $+0.132$ & $-0.149$ & Transformer  & DFDC \\
    DeiT-B/16   & $+0.007$ & $-0.095$ & Transformer  & DFDC \\
    EffNet-B0   & $+0.196$ & $-0.006$ & EfficientNet & DFDC \\
    MaxViT-B    & $+0.249$ & $-0.103$ & Hybrid       & DFDC \\
    Xception    & $+0.151$ & $-0.023$ & CNN          & DFDC \\
    \midrule
    ResNet50    & $+0.145$ & $+0.036$ & CNN          & None \\
    ResNet101   & $+0.178$ & $+0.008$ & CNN          & None \\
    EffNet-B4   & $+0.150$ & $+0.044$ & EfficientNet & None \\
    EffNetV2-S  & $+0.149$ & $+0.026$ & EfficientNet & None \\
    \bottomrule
  \end{tabular}
\end{table}

This architectural pattern has a practical interpretation: hybrid
architectures (Swin-B, ConvNeXt-B, MaxViT-B) rely on global attention
patterns that differ drastically between FF{$+\!+$} and in-the-wild
datasets, causing their learned uncertainty representations to invert
most severely. Traditional CNNs and EfficientNets, which exploit
local texture artifacts, produce more stable uncertainty estimates
under domain shift; their DFDC inversions, where present, are
small-magnitude ($|\rho| < 0.03$) compared to the Transformer and
most hybrid families ($|\rho| \gtrsim 0.07$, the exception being
ConvNeXt-B at $|\rho| = 0.042$). Forensic practitioners should
prefer CNN architectures (in particular ResNet50, ResNet101) or the
larger EfficientNets when cross-dataset reliability is critical and
no domain adaptation is possible.
 

%% file: sections/5-discussion.tex
\section{Discussion}
\label{sec:discussion}
 
\subsection{Forensic Implications}
 
For practitioners deploying deepfake detectors in forensic contexts,
our expanded eleven-architecture evaluation provides refined guidance.

\begin{figure*}[t]
  \centering
  \includegraphics[width=0.9\textwidth]{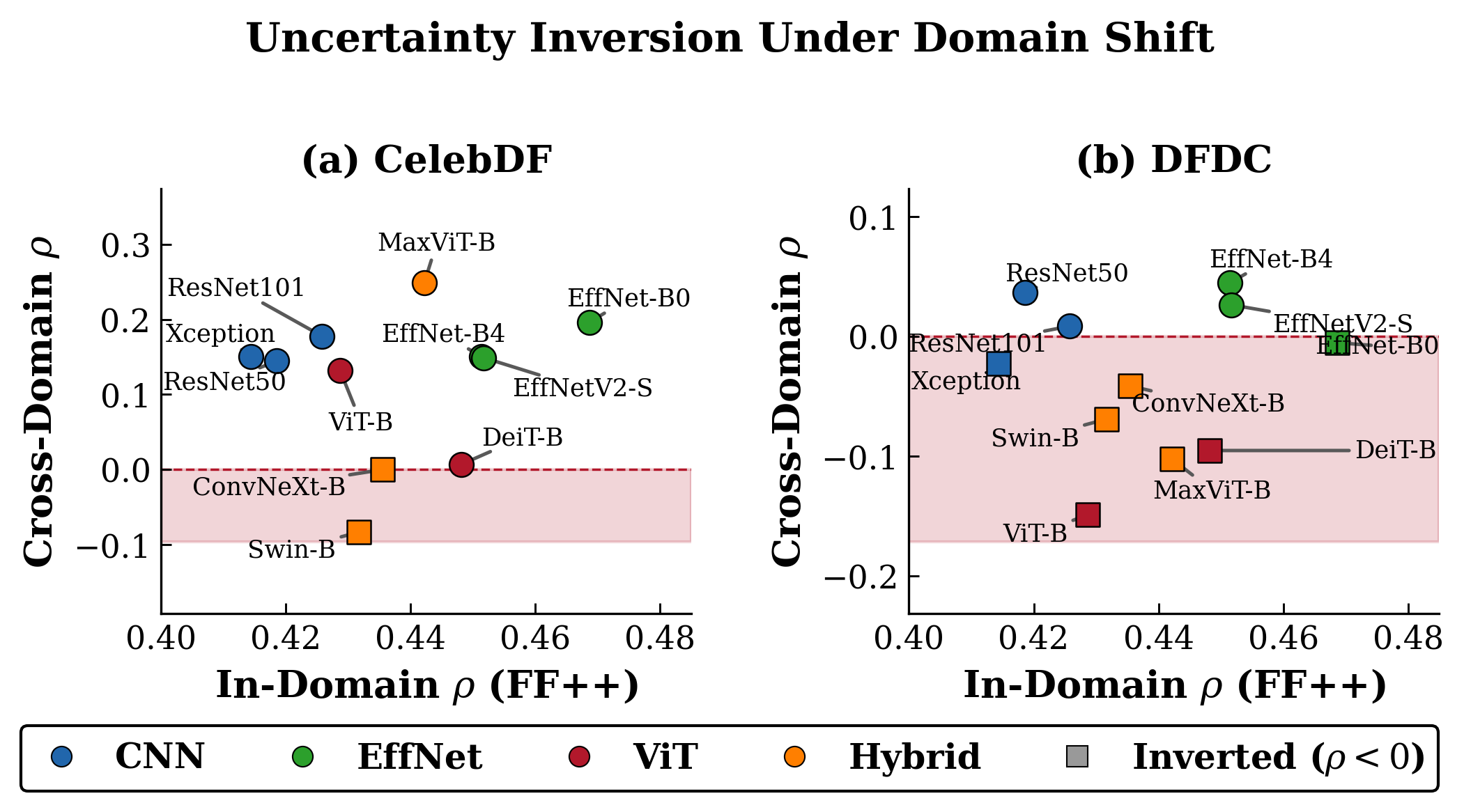}
  \vspace{-0.35cm}
  \caption{Uncertainty inversion analysis: in-domain vs.\ cross-domain
  correlation for each architecture. Points below the horizontal
  dashed line ($\rho = 0$) indicate uncertainty inversion --- models
  become \emph{more confident on errors} under distribution shift.
  (a)~CelebDF: Swin-B and ConvNeXt-B (orange squares) exhibit
  inversion, while CNNs and EfficientNets maintain positive
  correlation. (b)~DFDC: Seven architectures exhibit inversion,
  with ViT-B showing the strongest ($\rho = -0.149$). Xception is
  now also inverted on DFDC ($\rho = -0.023$), though at smaller
  magnitude than the Transformer/Hybrid families. The pink shaded
  region highlights the forensically dangerous inversion zone: high
  in-domain confidence paired with negative cross-domain correlation.}
  \label{fig:inversion_analysis}
\end{figure*}

\textbf{For cross-domain deployment} (unknown source distributions):
COF or L1-COF warrant consideration over non-linear alternatives. Under
matched training-data protocol (Table~\ref{tab:cof_vs_rf}), COF achieves $62\%$ higher cross-domain correlation on CelebDF than
Random Forest (mean $\rho = 0.116$ vs.\ $0.071$) despite slightly
lower in-domain $\rho$ ($0.438$ vs.\ $0.463$, a $5.7\%$ gap). This advantage
is consistent with the capacity-control argument of
Section~\ref{ssec:theory}: restricting $\w$ to the simplex tightens
the generalization bound under covariate shift.

\textbf{For controlled-distribution deployment} (known generation
pipeline): Non-linear methods (RF, Neural, Logistic) may be used for
maximum in-domain performance, but practitioners should validate on
held-out data from the expected deployment distribution.

\textbf{Architecture selection:} ResNet50, ResNet101, EfficientNet-B4,
and EfficientNetV2-S avoid uncertainty inversion on both CelebDF and
DFDC. All hybrid architectures (Swin-B, ConvNeXt-B, MaxViT-B) exhibit
inversion on at least one external dataset, as do both Transformers
(ViT-B, DeiT-B); Xception shows mild DFDC inversion ($\rho = -0.023$)
but remains stable on CelebDF. For deployments requiring cross-domain
reliability, the residual-CNN and larger EfficientNet families are
the safer choice.

\textbf{Conformal prediction} should always be included: it receives
25--46\% of learned weight across all architectures and is the most
critical source---removing it causes 4--34\% degradation. For zero-cost
uncertainty, SC-Weight achieves 98.4\% of COF's performance with no
optimization.

 \textbf{Distributional uncertainty for cross-domain deployment:}
Although distributional uncertainty has the lowest in-domain correlation
($\rho = 0.086$), it is the only source that maintains its correlation
cross-domain---and the only one that improves on CelebDF
(Table~\ref{tab:source_stability}).
Future domain-adaptive UQ systems should prioritize feature-space
distance measures over prediction-derived signals.

\subsection{The Reliability Gap: Dataset-Specificity of UQ}

The cross-dataset collapse (mean $\rho = 0.438$ in-domain to
$0.041$ externally, averaged across CelebDF and DFDC) is the
central finding for the TIFS community: current UQ methods,
regardless of sophistication, produce uncertainty estimates that
are fundamentally dataset-specific. This is not a COF limitation
--- all twelve methods including non-linear baselines exhibit the
same collapse pattern.

On DFDC, all eleven architectures classify 100\% of samples as
fake (prediction collapse), producing zero prediction variance.
Uncertainty sources derived from prediction output (epistemic,
aleatoric, calibration, conformal) become entirely degenerate;
only distributional uncertainty retains signal by operating on
feature-space distance independent of prediction output. This
collapse is distinct from accuracy degradation: a model achieving
70\% accuracy by predicting everything as fake still produces
completely uninformative uncertainty estimates.

The root cause lies in the per-source estimates themselves
(Table~\ref{tab:source_stability}): all four prediction-derived
sources collapse $>$90\% cross-domain, while distributional
uncertainty --- the weakest in-domain source ($\rho = 0.086$)
--- is the only source that does not collapse ($-2.9\%$ mean vs.\
$>$90\% for the rest) and in fact \emph{exceeds}
its in-domain value on CelebDF ($\rho = 0.155$). Feature-space
distance is inherently robust to the calibration failures that
cripple prediction-derived sources under distribution shift,
identifying domain-adaptive UQ as the critical open research
problem.

\subsection{Limitations and Failure Modes}
 
COF has three identifiable failure modes. First, near-zero error
rate ($< 2\%$) makes $\e$ approximately constant and Pearson correlation
undefined; practitioners should monitor error rates and fall back to
entropy-based scoring when the detector is near-perfect. Second, on small
validation sets ($N < 200$), correlation estimates are noisy and weight
optimization is unreliable; SC-Weight is preferable in such cases.
Third, the simplex constraint assumes uncertainty sources have comparable
reliability ranges after normalization, which may not hold if one source
has extreme outliers.
 
The COF objective is linear in $\w$ and optimizes Pearson correlation.
While we verify that Spearman $\rho_s$ and AUROC preserve all
qualitative conclusions (Table~\ref{tab:forensic_utility}), a
rank-based objective such as $\rho_s$ or concordance-index maximization
may be preferable when the uncertainty--error relationship is
strongly non-monotonic. Information-theoretic objectives (e.g.,
mutual information with $\e$) are an orthogonal direction for future
work.

Fourth, COF's linear fusion constrains in-domain performance relative
to non-linear alternatives. Our results show that Random Forest achieves
approximately $5\!-\!6\%$ higher in-domain correlation by capturing
source interactions that the simplex constraint prevents. However,
this limitation is simultaneously a strength: the same linearity that
limits in-domain expressiveness is empirically associated with better
cross-domain generalization, suggesting a regularization effect,
making COF more suitable for the forensic deployment scenarios where
domain shift is the primary concern.

Fifth, the DFDC prediction collapse (all models predicting 100\% fake)
indicates that the trained models' decision boundaries are entirely
FF++-specific. Future work should investigate whether domain adaptation
at the model level (not just the uncertainty level) is necessary before
uncertainty fusion can be meaningful on highly shifted datasets.
 
\subsection{Future Directions}
 
The cross-dataset collapse identifies domain-adaptive uncertainty fusion
as the central open challenge. Promising directions include:
meta-learning-based COF that rapidly adapts weights to new
domains using small calibration sets~\cite{nguyen2025think}; adversarial domain alignment of uncertainty
representations; attention-pattern regularization for Transformer
architectures to reduce cross-dataset inversion; and online weight
adaptation as distribution shift is detected during deployment. 
 

%% file: sections/6-conclusions.tex
\section{Conclusion}
\label{sec:conclusion}
 
We introduced Correlation-Optimized Fusion (COF), a post-hoc,
architecture-adaptive framework for reliable uncertainty
quantification in deepfake detection. Evaluation across eleven architectures spanning CNN,
EfficientNet, Transformer, and Hybrid families reveals
three central findings.
 
First, COF provides a principled, interpretable reliability signal
that generalizes better than non-linear alternatives under
distribution shift. While Random Forest achieves marginally higher
in-domain $\rho$ ($0.463$ vs.\ COF's $0.438$, a $5.7\%$ gap) under
matched train/test protocol, it degrades $85\%$ cross-domain to
$\rho = 0.071$, whereas COF retains substantially more signal
($74\%$ drop to $\rho = 0.116$). COF outperforms RF on
CelebDF in 9 of 11 architectures, with up to $7.3\times$ higher
cross-domain correlation (MaxViT-B: $\rho = 0.249$ vs.\ $0.034$).
The simplex constraint empirically acts as a form of regularization
against cross-domain overfitting, positioning COF as a practical option when capacity-controlled fusion is desired and cross-domain reliability is primary concern.
 
Second, conformal prediction is the single most critical uncertainty
source, receiving 25--46\% of learned weight and causing 4--34\%
degradation when removed across all eleven architectures. Epistemic
and calibration sources receive zero weight in nearly all
architectures, suggesting that MC Dropout variance and
temperature-scaled confidence are redundant with conformal and
aleatoric signals.
 
Third, cross-dataset evaluation exposes catastrophic failure across
all methods: in-domain correlations of 0.41--0.47 collapse to
near-zero externally (mean degradation 90.7\%), with complete
prediction collapse on DFDC. Seven of eleven architectures exhibit
uncertainty inversion. This identifies domain-adaptive UQ as the
central open problem for reliable deepfake detection in the wild.